\documentclass[runningheads]{llncs}
\pdfoutput=1
\usepackage{graphicx}
\usepackage{amsmath,amssymb} 
\usepackage{color}
\usepackage{float}
\usepackage{subfig}
\usepackage{wrapfig}
\usepackage{sidecap}
\usepackage{adjustbox}
\usepackage[width=122mm,left=12mm,paperwidth=146mm,height=193mm,top=12mm,paperheight=217mm]{geometry}
\usepackage{algorithm,algcompatible,amssymb,amsmath}
\renewcommand{\COMMENT}[2][.5\linewidth]{%
  \leavevmode\hfill\makebox[#1][l]{//~#2}}
\algnewcommand\algorithmicto{\textbf{to}}
\algnewcommand\RETURN{\State \textbf{return} }
\begin{document}
\pagestyle{headings}
\mainmatter
\def\ECCV16SubNumber{925}  

\title{Unsupervised Domain Adaptation in the Wild: Dealing with Asymmetric Label Sets} 

\title{Unsupervised Domain Adaptation in the Wild}


\author{Ayush Mittal\inst{1} \and Anant Raj \inst{2} \and Vinay P. Namboodiri \inst{1} \and Tinne Tuytelaars \inst{3} }

\institute{Department of Computer Science and Engineering, IIT Kanpur, India\\
\and
Max Planck Institute for Intelligent Systems, Tuebingen, Germany \\
\and 
ESAT, PSI-VISICS, KU Leuven, Heverlee, Belgium
\email{\{ayushmi,vinaypn\}@iitk.ac.in,  anant.raj@tuebingen.mpg.de,  tinne.tuytelaars@esat.kuleuven.be} }

\maketitle

\begin{abstract}
The goal of domain adaptation is to adapt models learned on a source domain to a particular target domain. Most methods for unsupervised domain adaptation proposed in the literature to date, assume that the set of classes present in the target domain is identical to the set of classes present in the source domain. This is a restrictive assumption that limits the practical applicability of unsupervised domain adaptation techniques in real world settings (``in the wild"). Therefore, we relax this constraint and propose a technique that allows the set of target classes to be a subset of the source classes. This way, large publicly available annotated datasets with a wide variety of classes can be used as source, even if the actual set of classes in target can be more limited and, maybe most importantly, unknown beforehand.

To this end, we propose an algorithm that orders a set of source subspaces that are relevant to the target classification problem. Our method then chooses a restricted set from this ordered set of source subspaces. As an extension, even starting from multiple source datasets with varied sets of categories, this method automatically selects an appropriate subset of source categories relevant to a target dataset. Empirical analysis on a number of source and target domain datasets shows that restricting the source subspace to only a subset of categories does indeed substantially improve the eventual target classification accuracy over the baseline that considers all source classes. 

\end{abstract}
\vspace{-1cm}
\section{Introduction}
\vspace{-0.2cm}
Suppose we have access to a state-of-the-art classifier trained with thousands of categories as in Imagenet and would like our household robot to use it to recognize objects. Unfortunately, as the distribution of samples in our house varies from that used in the Imagenet dataset, our state-of-the-art classifier will not perform that well. This is generally the case, as machine learning algorithms usually assume that the training and test instances come from the same distribution. This assumption does not hold for several real world computer vision problems. The distribution of image samples in our training set, termed {\em Source} domain, is different from the distribution of samples at test time, termed {\em Target} domain. This shift in the underlying data distribution is called {\em domain shift}.
Visual domain shift between two domains can be caused, among others, by differences in resolution, viewpoints, illumination, clutter and backgrounds. Domain shift problems are very pertinent in computer vision tasks because of the fact that each dataset has its own bias. That is why results from one dataset cannot easily be transferred to  another, as shown by Torralba and Efros \cite{torralba2011unbiased}. A number of domain adaptation techniques have been proposed in the literature \cite{gopalan2011domain,gong2012geodesic,fernando2014subspace,baktashmotlagh2013unsupervised,patel2015visual} that allow for adapting a source classifier to a target domain. However, all these methods assume that we a priori choose the categories in our source dataset that are applicable in our target dataset. This is not always applicable, as illustrated by our household robot example, where we would have to carefully modify the source dataset for each robot and for each house in which we want to deploy the robot. We would rather prefer an automatic solution that carefully chooses the subset of source categories best suited for a particular target dataset. This is the problem that we address in this paper.

Our main contribution is a general domain adaption algorithm that works even when the original label set of source and target data are different. We achieve this by iteratively modifying the source subspace. In each step we choose one more category from the source dataset to better adapt to the target dataset. This requires us also to define a stopping criterion that indicates when it is no longer useful to add a new category from the source domain. The method proposed relies on projection error between source and target domains to obtain the ordering of source category classes
and evaluates local and global minima based stopping criteria. 
Through this paper we make the following contributions:
\begin{itemize}
\item We consider the case of adapting a large/varied source dataset to a target domain dataset. This is achieved by identifying an ordered set of source categories.
\item We evaluate local and global minima based stopping criteria and show that choosing a restricted set of source categories aids adaptation to the target dataset.
\item We also further consider the case where multiple source datasets are present and the optimal combination of classes from these is selected and adapted to a particular target dataset.
\end{itemize}

The remainder of the paper is organized as follows. In the next section we discuss the related work, and in section \ref{background} we recapitulate some background material. In section \ref{approach}, the proposed method is discussed in detail. In section \ref{multisource} we consider the setting where multiple source datasets are adapted to a particular target dataset.
Experimental evaluation is presented in section \ref{experiments}, including the performance evaluation and a detailed analysis. We conclude in section \ref{conclusion}.

\vspace{-0.2cm}
\section{Related Work} \label{related}
\vspace{-0.2cm}
Though domain adaptation is a relatively new topic within the computer vision community, it has long been studied in machine learning, speech processing and natural language processing communities \cite{margolis2011literature,glorot2011domain}. We refer to~\cite{patel2015visual} for a recent review on visual domain adaptation. There are basically two classes of domain adaptation algorithms: semi-supervised and unsupervised domain adaptation. Both these classes assume the source domain is a labeled dataset. Semi-supervised domain adaptation requires some labeled target examples during learning, whereas unsupervised domain adaptation does not require any labeled target examples. In this work, we will focus on the unsupervised setting.
\vspace{-0.2cm}
\paragraph{\bf Deep adaptation} With the availability of convolutional neural  networks (CNNs), a few methods specific to CNNs have been proposed to adapt the deep network parameters. In \cite{oquab2014learning}, Oquab {\em et al.} showed that an image representation learned by CNN on a large dataset can be transferred for other tasks by fine tuning the deep network. These however assume availability of labeled data in the target dataset. We work in the unsupervised setting. In~\cite{long2015learning},the hidden representations of all the task-specific layers are embedded to a RKHS for explicit matching of the mean embeddings of different domain distributions. In the present work, we focus on the asymmetric label setting and can benefit from such work that investigates unsupervised domain adaptation of deep networks.

\vspace*{-0.2cm}
\paragraph{\bf Subspace-based methods} 
In our work we mainly work with subspace based methods using deeply learned feature representations. Subspace-based methods~\cite{gopalan2011domain,gong2012geodesic,fernando2014subspace} have shown promising results for unsupervised domain adaptation problems. In~\cite{gopalan2011domain}, source and target subspaces are represented as separate points on a Grassmann manifold. The data is then projected onto a set of in-between subspaces along the geodesic path, so as to obtain a domain invariant representation. One main issue with this approach is that we don't know how many in between subspaces need to be sampled along the path. To overcome this challenge, Gong et al. \cite{gong2012geodesic} proposed a geodesic flow kernel based approach that captures the incremental change from source to target subspace along the geodesic path.  
In \cite{fernando2014subspace}, Fernando et al. propose to learn a transformation by minimizing the Frobenius norm of the difference function which directly aligns source and target subspaces. 
Anant et al. \cite{raj2015mind} 
consider domain adaptation in a hierarchical setting and show that using different subspaces at different levels of hierarchy improves adaptation. 
Another direction of particular relevance to our work is the idea of continuously evolving domains as proposed in Hoffman et al.  \cite{hoffman2014continuous}. The key idea in their approach is to consider continuous temporal evolution of the target domain. We also use projection errors as our criterion for searching for subspaces in this work. However, none of the methods mentioned above try to address unsupervised domain adaptation task with an asymmetric distribution of labels. With the availability of large labeled source datasets, investigation of this setting merits attention and is carried out in this paper.


\section{Background} \label{background} 

In this section we formally define the learning task in domain adaptation and give some background about the Subspace Alignment approach for domain adaptation on which we build up our method.

\subsection{Learning Task} \label{learningtask}

We have a set $\mathcal{S}$ of labeled source data and a set $\mathcal{T}$ of unlabeled target data. Both $\mathcal{S}$ and $\mathcal{T}$ are $D$ dimensional. The source label set is denoted by $C_s = \{C_{s1},\dots,C_{sm}\}$ and the target label set by $C_t= \{C_{t1},\dots,C_{tn}\}$, where $m$ is the number of classes in the source domain and $n$ is the number of classes in the target domain.
Generally, domain adaptation algorithms assume source and target domains have the same label set i.e. $C_s = C_t$. Source $\mathcal{S}$ is drawn from a source distribution $\mathcal{D}_s$ and target $\mathcal{T}$ is drawn from target distribution $\mathcal{D}_t$. In general $\mathcal{D}_s \neq \mathcal{D}_t$. The aim in domain adaptation is to generalize a model learned on source domain $\mathcal{D}_s$ to target domain $\mathcal{D}_t$.

\subsection{Subspace Alignment}
The subspace alignment approach \cite{fernando2014subspace} for domain adaptation first projects the source data $(\mathcal{S})$ and target data $(\mathcal{T})$ to their respective $d$-dimensional subspaces using Principal Component Analysis. Let us denote these subspaces by $X_s$ for source and $X_t$ for target ($X_s,X_t\in \mathbb{R}^{D\times d}$). Then subspace alignment directly aligns the two subspaces $X_s$ and $X_t$ using a linear transformation function. Let us denote the transformation matrix by $M$. To learn $M$, subspace alignment minimizes following Bergmann divergence:

$$F(M) = ||X_sM-X_t||_F^2$$
$$M^* = argmin_M (F(M))$$
where $\| \cdot \|_F$  denotes Frobenius norm. Using the fact that $X_s$ and $X_t$ are intrinsically regularized and Frobenius norm is invariant to orthonormal operations, it can be easily shown that the solution for the given optimization problem is $M^*=X_s'X_t$ \cite{fernando2014subspace}. We denote a matrix transpose operation by $'$. The target aligned source subspace is then given by $U_{s} = X_sX_s'X_t$, ($U_s\in \mathbb{R}^{D\times d}$). To compare source data point $\mathbf{y}_s$ with a target data point $\mathbf{y}_t$ ($\mathbf{y}_s,\mathbf{y}_t \in \mathbb{R}^{D\times 1}$), the Subspace Alignment approach uses the following similarity measure: 

$$Sim(\mathbf{y}_s,\mathbf{y}_t) = \mathbf{y}_s'A\mathbf{y_t}$$

where $A = X_sMX_t'$. This similarity measure is then used as a kernel to train an SVM classifier.

\section{Approach} \label{approach}

In this section we introduce our algorithm to adapt for asymmetrically distributed label sets between source and target class. The algorithm can be divided into two steps. The first step iteratively expands the source subspace to discover target categories by minimizing projection errors and   the second step uses these categories to train a source classifier in the expanded subspace. 

\begin{figure}[H]
\centering
\includegraphics[width=0.9\textwidth]{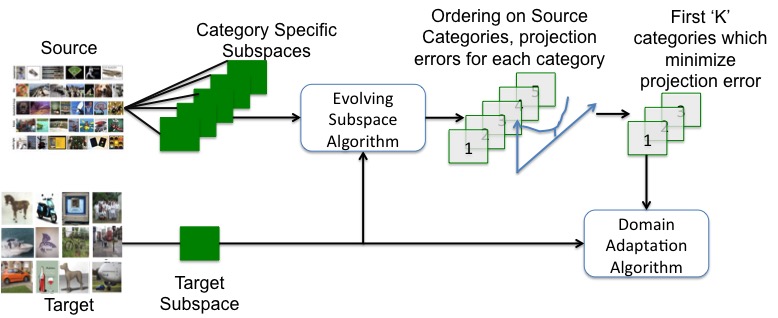}
\caption{Brief overview of evolving subspace algorithm.}
\label{fig:algorithm}
\end{figure}

In Figure \ref{fig:algorithm}, we illustrate the steps involved in the proposed approach. Evolving subspace algorithm evolves source subspace by iteratively adding categories that minimizes the projection error between source and target subspace. This iterative algorithm gives an ordering on the source categories. It also outputs the projection error after each iteration of the algorithm. We first select $K$ categories from this ordering which minimize the projection error. We can then use any domain adaptation algorithm to learn a classifier using these selected source categories to adapt to target.  

We first define two different projection error criteria that are used later  to order the categories from source. Then we describe the two steps of our algorithm.

\subsection{Projection Errors} \label{proj}

Projection Error is a key part of our algorithm as it decides which categories minimize the discrepancy between source and target subspaces. We introduce two different projection errors. Given the $d$-dimensional source subspace $X_s$ and the $d$-dimensional target subspace $X_t$ ($X_s,X_t \in \mathbb{R}^{D\times d}$), let us denote by $U_s$ the target aligned source subspace i.e. $U_s = X_sX_s'X_t (\in \mathbb{R}^{D\times d})$ \cite{fernando2014subspace}. Similar to the subspace alignment criterion described in Section \ref{background}, our first projection error is the following Bregmann matrix divergence (we call it Subspace Alignment Error):
$$
Subspace Alignment Error = ||U_s - X_t||_F
$$
where $F$ denotes Frobenius norm. This subspace alignment error is a subspace based criterion. Intuitively, domains may share principal components. Subspace alignment aligns principle components which makes adaptation possible. If a category is present in both source and target then corresponding category specific subspaces for source and target will share principal components and subspace alignment error will be small for such categories. This makes subspace alignment error a good projection error for our problem. Our second criterion is Reprojection Error introduced by Hoffman et al. \cite{hoffman2014continuous}. Reprojection error is a sampling based criterion which uses source examples instead of source subspace for computing projection error. Hoffman et al. \cite{hoffman2014continuous} compute the Reprojection Error of the source data. We instead use the following Reprojection error of target data on target aligned source subspace $(U_s)$:
$$
Reprojection Error = ||\mathcal{T} - (\mathcal{T}U_s)U_s'||_F
$$
where $\mathcal{T} (\in \mathbb{R}^{N_t\times D})$ is the target data. Reprojection error is a sampling based criterion. 
Intuitively, reprojection error tries to measure the discrepancy between original target data matrix and reconstructed target data matrix after projecting them onto target aligned source subspace. 
Algorithm \ref{algsa} and algorithm \ref{algrep} give the functions for computing Subspace Alignment Error and Reprojection Error respectively which we will use as a sub-routine in the next section.
\begin{algorithm}[H]
  \caption{Computing the Subspace Alignment Error\label{algsa}}
  \begin{algorithmic}
  	\State Input: Set of categories denoted by $C$
    \State $\mathcal{S_C} \leftarrow $  Source Examples with labels $\in C$
 	\State $\mathcal{T} \leftarrow $ Target Examples
    \State $X_t \leftarrow PCA(\mathcal{T})$			\COMMENT{Target Subspace}
   	\State $X_s \leftarrow PCA(S_C)$ \COMMENT{Source Subspace on given label set}
   	\State $U_s \leftarrow X_sX_s'X_t$ \COMMENT{Target aligned source subspace}
    \State Output: $||U_s-X_t||_F$ \COMMENT{Subspace Alignment Error}
  \end{algorithmic}
\end{algorithm}
\vspace*{-0.8cm}
\begin{algorithm}[H]
  \caption{Computing the Reprojection Error\label{algrep}}
  \begin{algorithmic}
  	\State Input: Set of categories denoted by $C$
    \State $\mathcal{S_C} \leftarrow $  Source Examples with labels $\in C$
 	\State $\mathcal{T} \leftarrow $ Target Examples
    \State $X_t \leftarrow PCA(\mathcal{T})$			\COMMENT{Target Subspace}
   	\State $X_s \leftarrow PCA(S_C)$ \COMMENT{Source Subspace on given label set}
   	\State $U_s \leftarrow X_sX_s'X_t$ \COMMENT{Target aligned source subspace}
    \State Output: $||\mathcal{T} - (\mathcal{T}U_s)U_s'||_F$ \COMMENT{Reprojection Error}
  \end{algorithmic}
\end{algorithm}
\subsection{Step 1: Evolving Subspace}

In this step we evolve the source subspace into the target subspace by iteratively adding source categories which minimizes the projection error of the subspace specific to these categories. Algorithm \ref{alg} gives the different steps involved. We start with an empty list of categories which we denote by $selectedCategories$. In each outer iteration of the algorithm we add a category to this list. The inner iteration uses Algorithm \ref{algsa},\ref{algrep} to compute the projection errors and selects a category with minimum error from the categories not selected so far. The algorithm finally outputs the list of categories added after each iteration ($selectedCategories$) and the list of  corresponding projection errors ($errors$) at the end of each iteration. Let us denote by $U_s$ the target aligned source subspace on the categories in the list $selectedCategories$. Intuitively, if we add a source category $C_{i}$ to the list $selectedCategories$ and the category $C_{i}$ is also present in the target label set, then adding $C_i$ will bring $U_s$ and the target subspace $X_t$ closer to each other. We use projection errors for deciding this closeness of the two subspaces. If projection error increases on adding category $C_{i}$ then the examples in $C_i$ are not similar to the target examples. This algorithm naturally gives us an ordering on the source categories: the order in which categories are selected based on minimizing the projection errors. Ideally, we expect categories present in target domain to occur first in the list $selectedCategories$. Hence we expect the projection error to first decrease after each iteration until all categories which are present in both source and target are added and then it must increase as we add other categories.
\begin{algorithm}[H]
  \caption{Evolving source subspace to learn target categories\label{alg}}
  \begin{algorithmic}
    \State $selectedCategories \leftarrow [\quad]$				\COMMENT{Empty List}
    \State $unselectedSet \leftarrow C_s$				\COMMENT{Set of categories in source}
    \State $errors \leftarrow [\quad]$ \COMMENT{Stores projection errors}
    \WHILE{$unselectedSet$ is not empty}
    	\State $minProjectionError \leftarrow$ INFINITE
        \State $selectedCategory \leftarrow -1$
        \FOR{Category $c$ in $unselectedSet$}
        	\State $error \leftarrow$ Projection Error on $selctedCategories \cup \{c\}$ (Algorithm \ref{algsa},\ref{algrep}) 
            \IF{$error < minProjectionError$}
            	\State $minProjectionError \leftarrow error$
                \State $selectedCategory \leftarrow c$
            \ENDIF
        \ENDFOR
        \State Remove $selectedCategory$ from $unselectedSet$
        \State Append $selectedCategory$ to $selectedCategories$
        \State Append $error$ for $selectedCategory$ to $errors$
    \ENDWHILE
   	\State Output: $selectedCategories,errors$
  \end{algorithmic}
\end{algorithm}
\subsection{Step 2: Training Source Classifier}

Step 1 gives us an ordering on the categories of the source domain. Let us denote this ordering by $selectedCategories = [C_{\alpha_1},C_{\alpha_2}\dots ,C_{\alpha_m}]$, where $m$ is the number of classes in the source label set. Let the projection errors after adding each category in step 1 be $errors = [E_1,E_2,\dots,E_m]$. As discussed before we expect values in $errors$ to first decrease and then increase. We propose to use this intuition and select the first $K$ categories of the list $selectedCategories$ such that $E_K (\in errors)$ is a minimum. We then use the categories $C_{train} = [C_{\alpha_1},C_{\alpha_2},\dots,C_{\alpha_K}]$ for training a source classifier in the target aligned subspace. Ideally, we should get a single global minimum in $errors$, but this is not true in practice and we get many local minima. Hence we propose to use two methods to select $K$. The first method is to select $K$ such that $E_K$ is the first local minimum in $errors$. The second method selects $K$ such that $E_K$ is the global minimum in $errors$. We discuss advantages and disadvantages of using local minima vs using global minima in section \ref{exp_evolution}. Algorithm \ref{alg2} outlines the steps described here.

\begin{algorithm}[H]
  \caption{Training Source Classifier\label{alg2}}
  \begin{algorithmic}
  	\State \textbf{Input}: \COMMENT{Output of Algorithm \ref{alg}}
    \State \quad $selectedCategories = [C_{\alpha_1},C_{\alpha_2}\dots ,C_{\alpha_m}]$
    \State \quad $errors = [E_1,E_2,\dots,E_m]$
    \State $K \leftarrow$ Index in $errors$ with Global minima/first local minima.
    \State $C_{train} \leftarrow [C_{\alpha_1},C_{\alpha_2}\dots ,C_{\alpha_K}]$ \COMMENT{Categories selected for training}
    \State $S_{train} \leftarrow $ Source examples with labels $\in C_{train}$
    \State $X_{S_{train}} \leftarrow PCA(S_{train})$ \COMMENT{Source Subspace}
    \State $X_t \leftarrow PCA(\mathcal{T})$ \COMMENT{Target Subspace}
    \State $U_S \leftarrow X_{S_{train}}X_{S_{train}}'X_t$ \COMMENT{Target Aligned source space}
    \State $trainingData \leftarrow S_{train}U_S$ \COMMENT{Source data in aligned space}
    \State $Classifier \leftarrow$ Train classifier on $trainingData$
    \State We then use this $Classifier$ to predict on target data in target space $X_t$.
    \State \textbf{Output}: $Classifier$
    \end{algorithmic} 
\end{algorithm}

\section{Multi-Source Domain Adaptation}
\label{multisource}
Simplicity of our algorithm allows us to extend our algorithm to multi-source domain adaptation where we have more than one source domains. The aim in multi-source domain adaptation is to generalize these multiple source domains to a single target domain.
This is particularly useful in practice, as a non-expert may not know which source dataset is to be preferred for his given target data (i.e., most similar to it).

\subsection{Problem Description} Given labeled source domains $\mathcal{S}_1,\mathcal{S}_2,\dots,\mathcal{S}_p$ and a target domain $\mathcal{T}$, the aim in multi-source domain adaptation is to generalize a model learned on these source domains to target domain $\mathcal{T}$.  Let $L = \{C_1,C_2,\dots,C_m\}$ denote the label set of source and target domains. A common approach for using multiple sources is to combine all the sources into a single source and then learn a classifier using single source domain adaptation. This approach ignores the difference between different source distributions. Another common approach is to train a classifier for each source domain separately and combine these classifiers.
In contrast, our algorithm automatically picks an optimal combination of categories from the different source domains. 

\subsection{Our Approach} We again use the idea of evolving source subspace by selecting categories which minimizes projection errors. For this problem, the same categories coming from different source domains are considered as different. Let us modify our label set $L$ to $L_* = \{C_{11},C_{12},\dots,C_{pm}\}$ where $C_{ij}$ represents category $C_{j}$ of source domain $\mathcal{S}_i$. We combine all the source domains and apply our single source adaptation algorithm given in the previous section (Section \ref{approach}) to this combined source domain with label set $L_{*}$. The evolving step (Algorithm \ref{alg}) of our approach will give an ordering on label set $L_{*}$, and the training step (Algorithm \ref{alg2}) will train a classifier on first $K$ categories of this ordering which minimizes overall projection error. Although our algorithm does not require the knowledge of the target label set, for the multi-source setting we evaluate with both known and unknown target label set. When using the target label set information, we add an additional constraint in our training step (Algorithm \ref{alg2}) that the category set $C_{train}$ used for training the source classifier should have each target category present at least once. If a target category $C_j$ is such that $C_{ij}$ is not in training set $C_{train}$ for all source domains $\mathcal{S}_i$ $(i \in 1\dots p)$ then we select the $C_{ij}$ which occurs first in the ordering. We give experimental results for multi-source domain adaptation in Section \ref{multiexp}.
\section{Experimental Results} \label{experiments}

This section is devoted to experimental results to validate our proposed algorithm. The performance evaluation has been done on numerous object classification datasets. SVM is used as the final classifier on transformed source data. Apart from comparing the empirical performances on several datsets, we also demonstrate the effect of varying the size of the target label set and analyze the effect of underestimating/overestimating the number of target categories on the adaptation. We intend to make our code available on acceptance.

\subsection{Datasets}

\textbf{Cross-Dataset Testbed \cite{tommasi2014testbed}}: Tommasi et al. \cite{tommasi2014testbed} have preprocessed several existing datasets for evaluating domain adaptation algorithms. They have proposed two setups: the dense set and the sparse set for cross-dataset analysis. Here we use the dense set for the evaluation of our algorithm. The dense set consists of four datasets Caltech256 \cite{griffin2007caltech}, Bing\cite{bergamo2010exploiting}, Sun\cite{xiao2010sun}, and ImageNet\cite{deng2009imagenet} which share 114 categories in themselves with 40 of the categories having more than 20 images per category. We use these 40 categories for the performance evaluation of the proposed algorithm.\\\\
\textbf{Office-Dataset \cite{saenko2010adapting}}: This datasets consists of three domains: Amazon (images obtained from Amazon website), DSLR (High resolution images obtained from DSLR camera), Webcam (Low resolution images obtained from a Webcam) each having 34 categories. As done in \cite{gong2012geodesic},\cite{fernando2014subspace} we use this dataset along with Caltech256 \cite{griffin2007caltech} which together have 10 overlapping classes for the performance evaluation in multi-source setting.

For all the domains, we use DeCAF \cite{donahue2013decaf} features extracted at layer 7 of the Convolutional Neural Network. These are the activation values of 7th layer of the Convolutional Neural Network trained on 1000 Imagenet classes. In the following sections we present evaluation of our algorithm on Bing\cite{bergamo2010exploiting} and Caltech256\cite{griffin2007caltech} datasets of dense setups. We give results for other datasets of dense setup in supplementary material.
\vspace{-0.3cm}
\subsection{Effect of Number of Source Categories on Adaptation} \label{exp:effectofnumberofcategories}

\begin{figure}[H]
\centering
\includegraphics[width=0.8\textwidth]{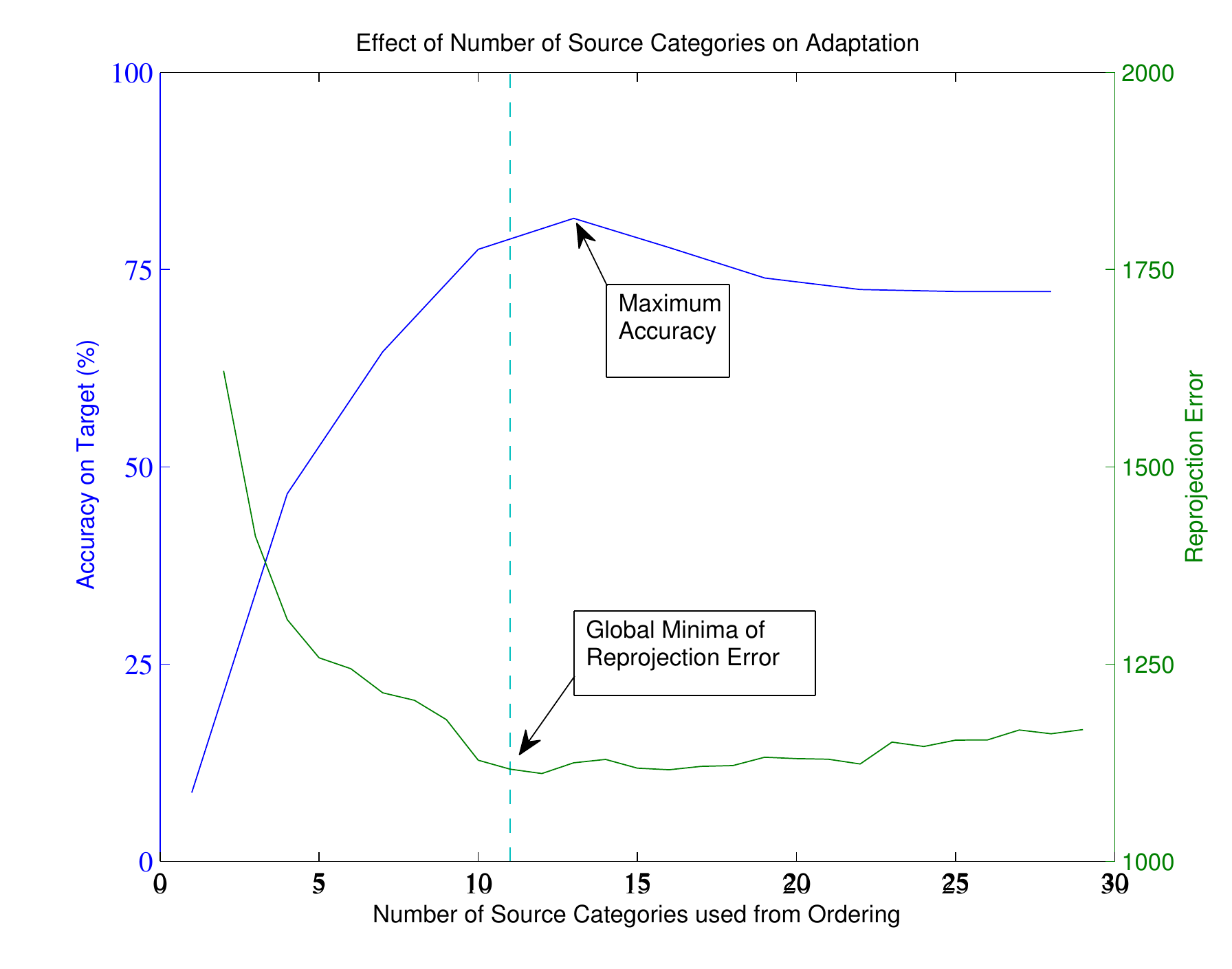}
\vspace{-0.4cm}
\caption{Performance of our adaptation algorithm (left y-axis) and reprojection error (right y-axis) as we change the number of source categories (x-axis) used for training the source classifier.}
\label{fig:effectofcategories}
\end{figure}

In this subsection, we demonstrate the effect of varying the number of source categories on the performance of domain adaptation. `Bing' is used as source dataset and `Caltech' as target.  We fix the size of the label set in the target data, setting it to 10. We apply our algorithm and greedily choose the source categories one by one in successive steps from the source dataset. The maximum no. of source categories which can be chosen is set to 30 categories. Figure \ref{fig:effectofcategories} shows the result for this experiment. We show both accuracy (blue) and reprojection errors (green) in Figure \ref{fig:effectofcategories}. The maximum accuracy is achieved when the number of selected source categories is 13 and the minimum rerprojection error is achieved when the number of selected source categories is 11.
\vspace{-0.5cm}
\vspace{-0.7cm}
It can be observed from the figure \ref{fig:effectofcategories} that for small number of source categories accuracy of the adapted classifier is low as a number of target categories are missing while learning the source subspace. It can also be clearly seen that having too many categories in source classifier which are not present in target doesn't help in improving classification accuracy because it increases the distance between source and target subspace. As mentioned earlier in section \ref{learningtask}, we expect the domain adaptation algorithm to perform best when source and target label sets are the same because under these conditions the distance between the subspaces would be the smallest. All the target categories are covered in first 13 source categories of the ordering which justifies the maximum value of accuracy for 13 source categories.
The plot also justifies our intuition for projection error as it first decreases, attains a minimum and then increases.
\vspace{-0.3cm}

\subsection{Analysis of Evolution Step}
\label{exp_evolution}
\vspace{-0.7cm}
\begin{figure}[ht]
  \centering
\subfloat[Using Reprojection Error and  First Local Minima]{\includegraphics[width=0.47\textwidth]{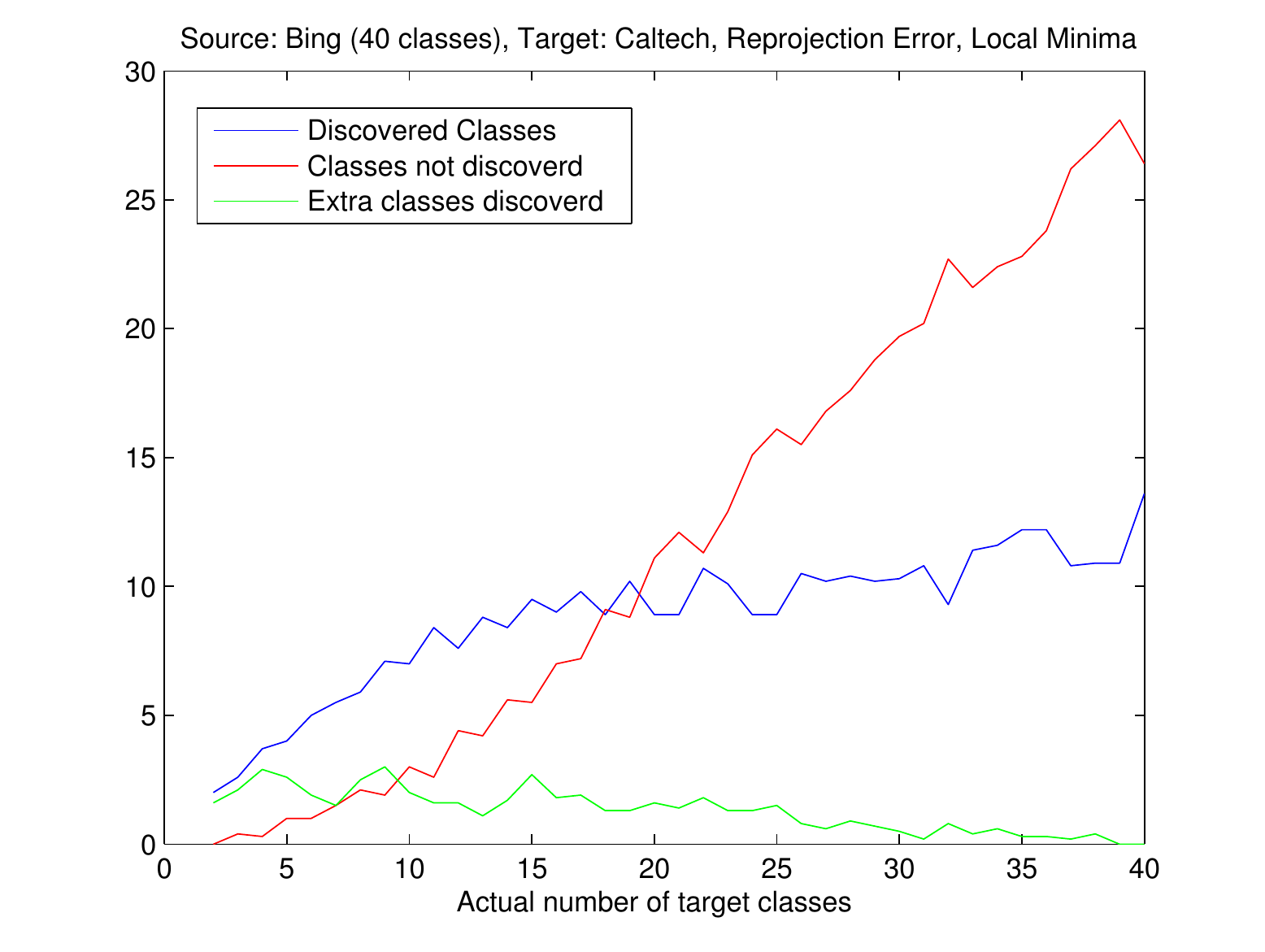}\label{fig:e1f1}} \hfill
  \subfloat[Using Subspace Alignment Error and First Local Minima]{\includegraphics[width=0.47\textwidth]{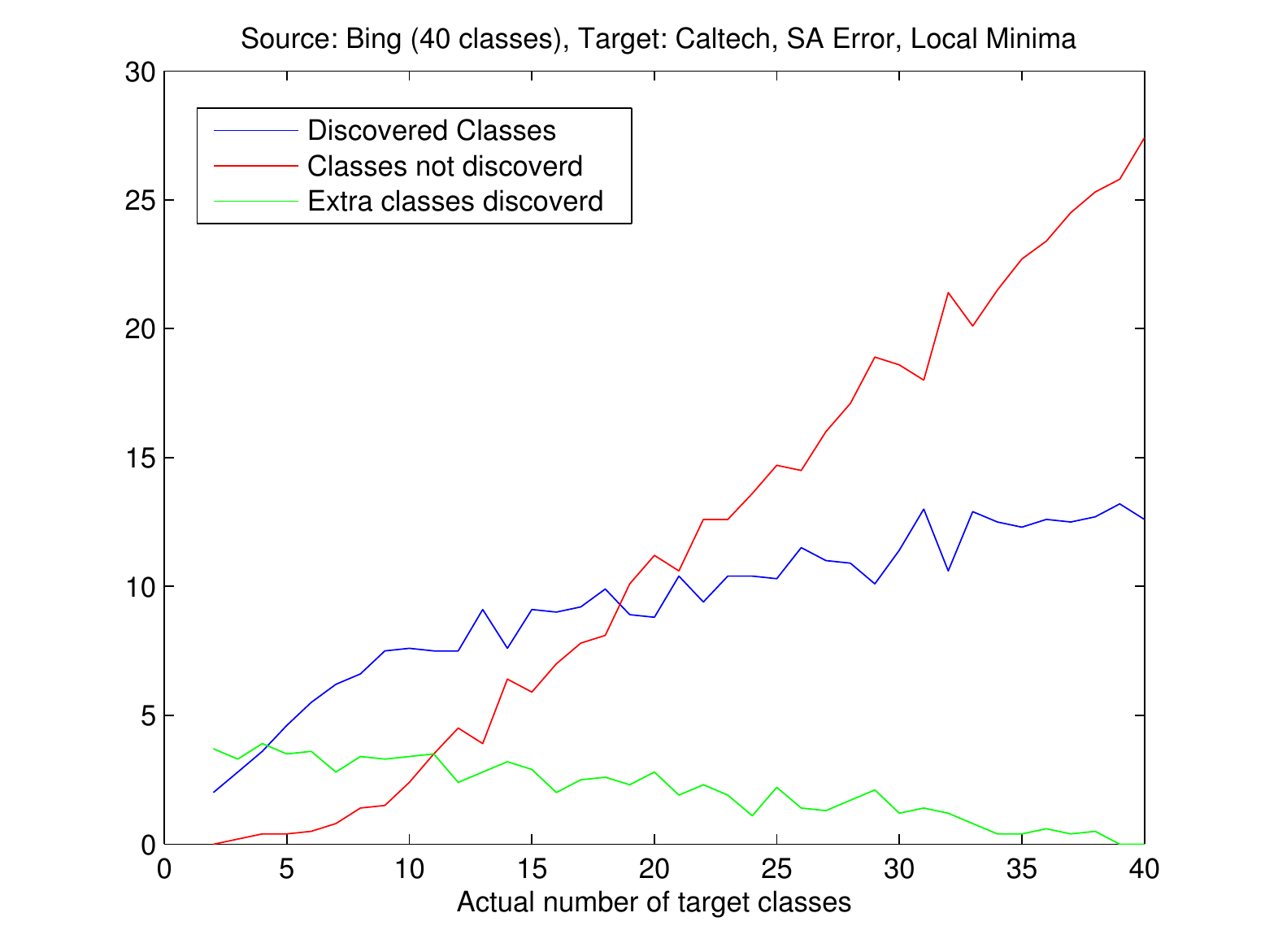}\label{fig:e1f3}}\\
  \subfloat[Using Reprojection Error and Global Minima]{\includegraphics[width=0.47\textwidth]{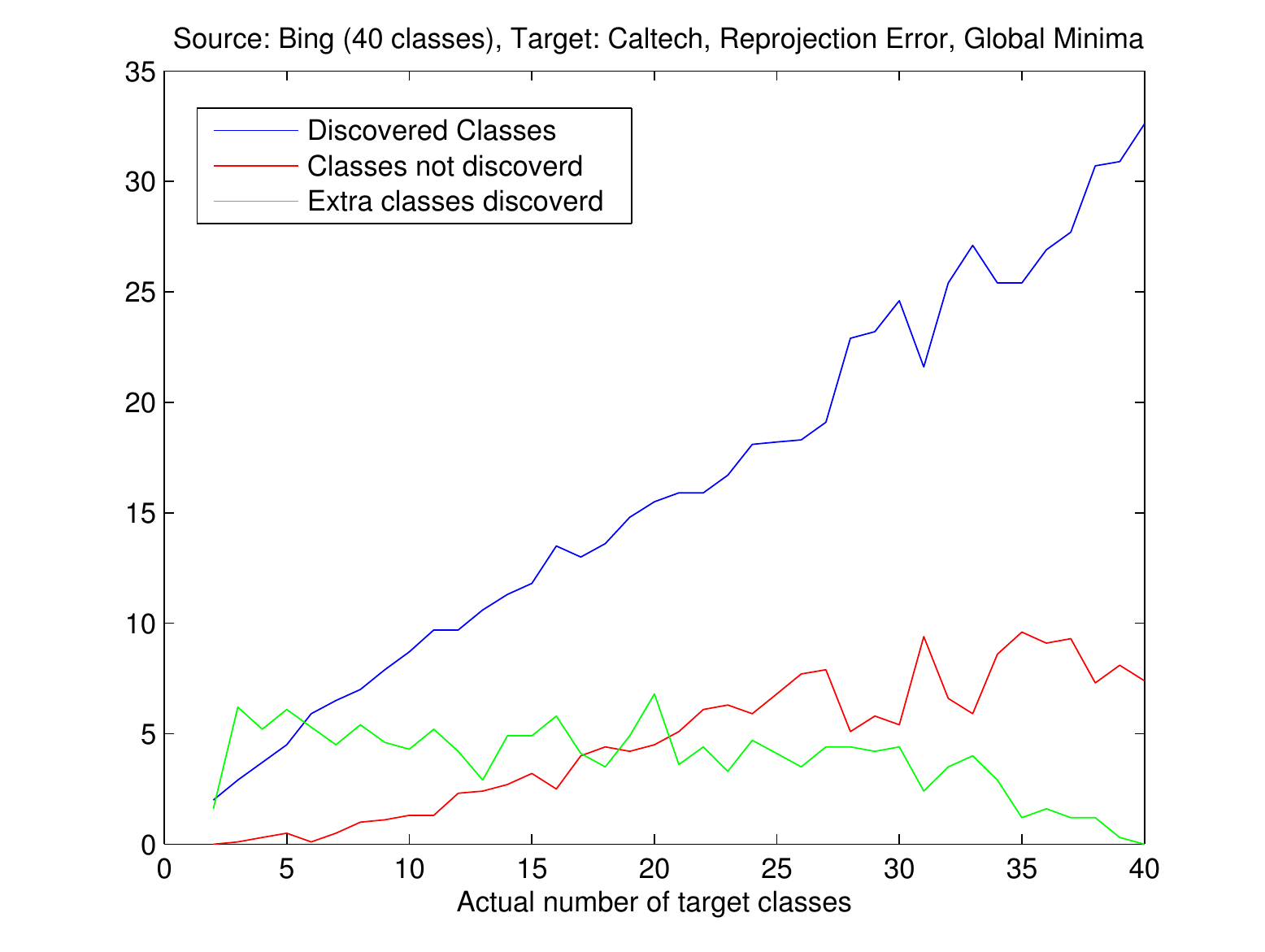}\label{fig:e1f2}}\hfill
  \subfloat[Using Subspace Alignment Error and Global Minima]{\includegraphics[width=0.47\textwidth]{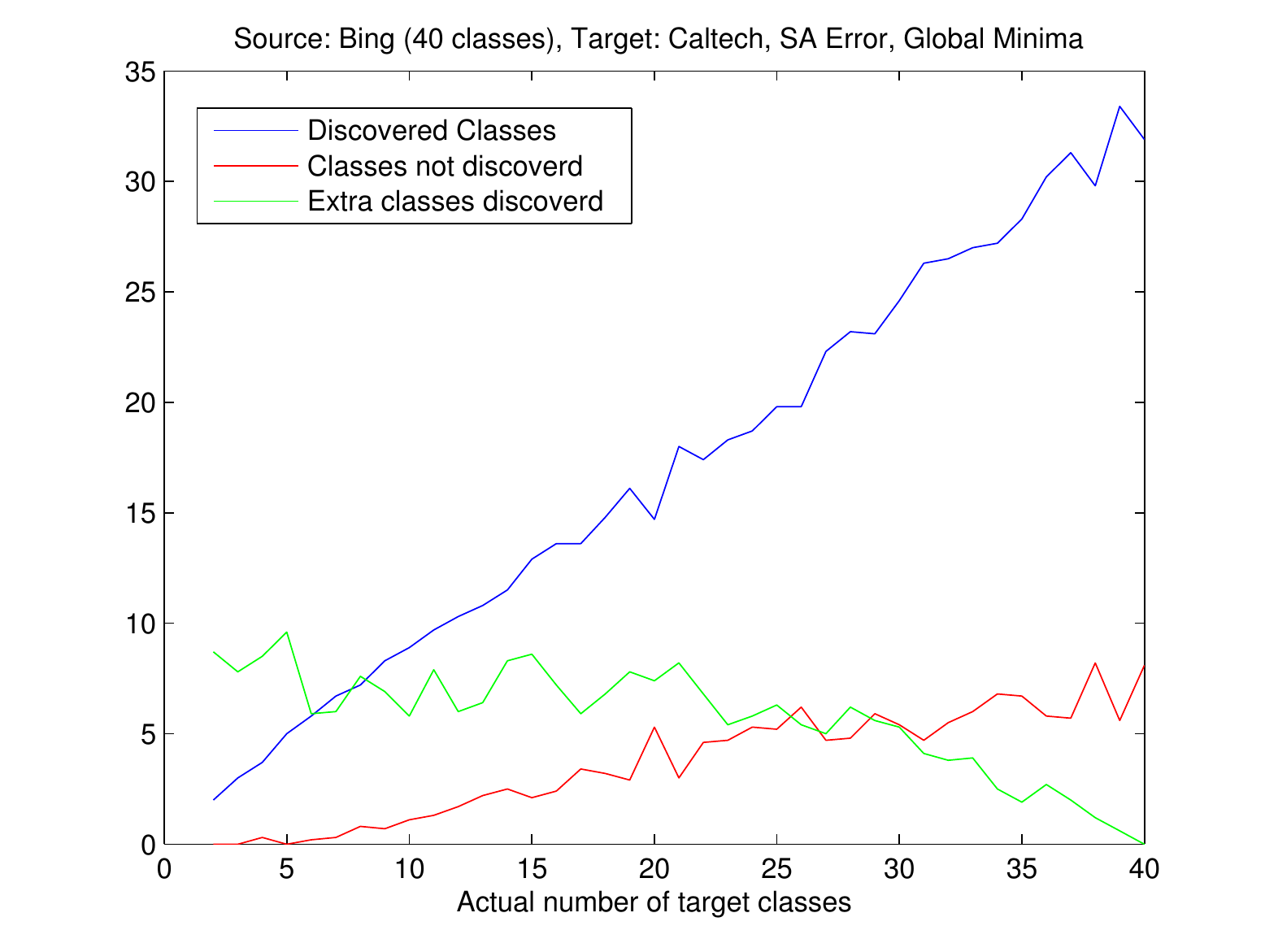}\label{fig:e1f4}}
  \vspace{-0.2cm}
  \caption{Performance of Evolution Step for different sizes of the target label set (x-axis). Blue: Number of target categories covered in final selected set $C_{train}$. Red: Number of target categories missed in final set. Green: Number of categories not present in target but selected in final set.}\label{fig:e1}
\end{figure}
\vspace{-0.5cm}
Our measure for analyzing the evolution step is based on the number of target categories present in the final list of selected categories by our algorithm. Let us assume that the source categories are  $(C_{src} = [C_{1},C_{2},\dots,C_{m}])$  and our algorithm selects  $(C_{train} = [C_{\alpha_1},C_{\alpha_2},\dots,C_{\alpha_K}])$ out of $C_{src}$ after the final evolution step of the algorithm. Figure \ref{fig:e1} shows the performance for Bing as source dataset and Caltech as target dataset. X-axis represents true number of categories in target dataset. For each point on x-axis we average over 10 permutation to get number of target categories correctly discovered, number of target categories missed and number of non-target categories in final selected set $C_{train}$. Ideally, we expect selected categories $C_{train}$ to be the label set of target domain. Hence, we expect a $x=y$ line in the plot. 

Reprojection Error (Figure \ref{fig:e1f1},\ref{fig:e1f2}) and Subspace Alignment Error (Figure \ref{fig:e1f3},\ref{fig:e1f4}) both show similar trends. For computing $K$ we propose to use either the first local minimum or the global minimum of projection errors. It can be clearly seen in Figure \ref{fig:e1f1},\ref{fig:e1f3} that the number of non-target categories discovered using the first local minimum is very small (Green curve) but it highly underestimates the number of target categories for large number of target classes (Blue vs Red curve). Hence, using the first local minimum discovers very small number of wrong classes but misses a lot of correct classes. Using the global minimum (Figure \ref{fig:e1f2},\ref{fig:e1f4} ) for selecting $K$ doesn't underestimate for large number of categories in target. However, for small number of categories using the global minimum overestimates. Although, we use global minimum for rest of the experiments, one can use first local minimum if one knows beforehand that the number of categories in target is small. As subspace alignment error and reprojection error show similar performance, we use reprojection error for the rest of the experiments. (For other pairs of source and target dataset in dense setup we give plots for this experiment in supplementary material).

\vspace{-0.3cm}
\subsection{Benchmark Comparison with SA as adaptation technique} 

\begin{figure}[H]
\centering
\includegraphics[width=10.5cm,height=7cm]{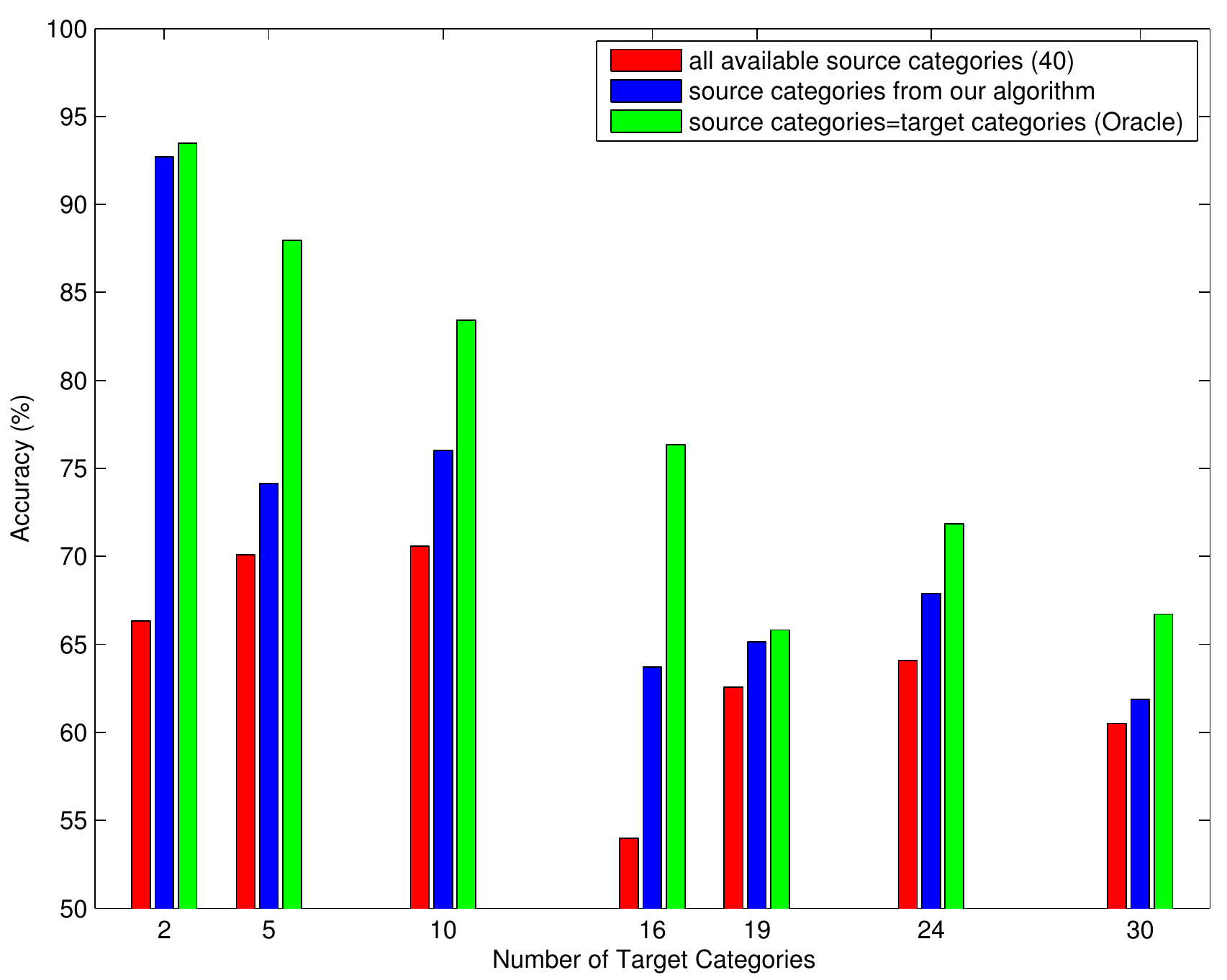}
\caption{Performance of our approach with subspace alignment based domain adaptation algorithm on Bing-Caltech dataset, with Bing as Source (40 categories) and Caltech as Target.}
\label{fig:benchmark}
\end{figure}

Next, we compare the performance of our algorithm with two extreme setups: First is the performance of domain adaptation algorithm while all available source categories are considered with respect to the target categories to learn the domain invariant representation and second is the performance of domain adaptation algorithm when only those source categories are considered which also belong to target categories (i.e., using an oracle). Here, we use subspace alignment algorithm~\cite{fernando2014subspace} for the final domain adaptation. 
Figure \ref{fig:benchmark} shows  the performance of our algorithm with subspace alignment based domain adaptation algorithm. Bing is used as source (40 categories), Caltech is used as target and for each point on x-axis accuracies are averaged over 10 experiments. As we have discussed in Section \ref{exp:effectofnumberofcategories}, we expect a domain adaptation algorithm to perform best when source and target categories have same label set. Green bars in Figure \ref{fig:benchmark} justify our intuition. Performance of categories given by our algorithm (blue bars) always outperform the case when we use all source categories for doing the domain adaptation and training the classifier (red bars).

\vspace{-0.2cm}
\subsection{Experiments with GFK as adaptation technique} 
\label{gfkexp}
It is important to note that once the categories from the source have been selected using the criteria proposed in section \ref{approach}, any  domain adaptation algorithm can be used afterwards. So far, we choose to use subspace alignment based domain adaptation approach to evaluate our algorithm. Here, we also provide  an experiment with geodesic flow kernel based approach \cite{gong2012geodesic} to validate our claim.  We use `Webcam' and `Amazon' domains of Office-Caltech dataset as target and source respectively. Results given in table \ref{table:gfk} show that GFK classifier trained using categories predicted by our algorithm outperforms the GFK classifier which uses all available source categories in almost all of the cases.
\vspace{-0.4cm}
\vspace{-1.2cm}
\begin{table}[H]
\begin{center}
\begin{adjustbox}{width=1\textwidth}
\begin{tabular}{|c|c|c|c|}
\hline
Number of Target & Source Categories = Target  & Available Source  & Predicted Target Categories  \\ 
Categories & Categories (Oracle) & Categories & (Our algorithm) \\
\hline
2 & 93.6701 & 64.0528 & \textbf{68.2893}\\
\hline
3 & 92.8030 & 64.6888 & \textbf{78.0935}\\
\hline
4 & 86.9177 & 64.3291 & \textbf{71.9271}\\
\hline
5 & 83.6397 & 68.1922 & \textbf{74.8842}\\
\hline
6 & 79.4589 & 68.7642 & \textbf{69.9612}\\
\hline
7 & 79.7467 & 71.3108 & \textbf{73.4498}\\
\hline
8 & 78.8076 & 72.0334 & \textbf{72.3195}\\
\hline
9 & 77.0064 & 73.4751 & \textbf{73.5467}\\
\hline
10 & 73.7835 &  \textbf{73.4263} & 71.7411\\
\hline
\end{tabular}
\end{adjustbox}
\vspace{0.2cm}
\caption{Performance of our algorithm with GFK (geodesic flow kernel) as adaptation technique. Source Datset:Webcam, Target Dataset:Amazon}
\label{table:gfk}
\end{center}
\end{table}
\vspace{-1.8cm}
\subsection{Multi-Source Domain Adaptation}\label{multiexp}
One more important application of our approach is in the case of multiple source domain adaptation. These settings occur very often in real world scenarios. Consider a case when you have lots of source domains available and have no information about the domain/domains which is/are most similar to the target. We apply our algorithm to choose the corresponding source categories from the best possible source domain to match with the respective target labels.
This section is devoted to the experiments on multi-source domain adaptation setting using our proposed approach. We use Office-Caltech dataset for this experiment. There are 10 overlapping categories in four domains of Office-Caltech dataset. We consider one of the four domains as target and rest three as source domains. Both source and target categories are assumed to be known. As discussed before we consider same category from different sources as different, we relabel source categories from 1 to 30 (10 for each source domain). Now, we run our algorithm with the extra constraint to select all the target categories at least once from some source domain. The experimental results for this setting has been reported in table \ref{table:multisource}. Along with the performance of our algorithm we list performance of single source domain adaptation with all three sources combined into a single source and also the performance of three single source domain adaptation classifier each with one of three sources. Note that our algorithm either outperforms or performs very comparable to all the other benchmarks mentioned above on Office-Caltech dataset. If we remove the constraint of selecting all the target categories at least once then we get the same results as that for the constrained case for DSLR, Amazon and Caltech as target datasets. For Webcam, number of target categories are underestimated and we get a minima at 7 instead of 10.

\vspace{-0.7cm}
\begin{table}[ht]
\begin{center}
\caption{Multi-Source domain adaptation results.}
\label{table:multisource}
\vspace{0.2cm}
\begin{adjustbox}{width=1\textwidth}
\begin{tabular}{|c|c|c|c|c|c|c|c|}
\hline
Target Domain &Source&Source&Source&Source&Source&Source&Number of\\
 &DSLR&Amazon&Webcam&Caltech&Rest three domains&Selected by our Algorithm&categories\\
\hline
DSLR &-& 86.00 & \textbf{97.33} & 78.00 & 70.67 & \textbf{97.33} & 10 \\
\hline
Amazon & 61.27 &-& 64.62 & 59.82 &62.05 &\textbf{74.44} &16 \\
\hline
Webcam & \textbf{82.91} & 72.95 &-&58.01 &66.55 &81.49 &10 \\
\hline
Caltech & 62.61 & 91.62 &70.87 &-&\textbf{91.74} & 90.85&17 \\
\hline
\end{tabular}
\end{adjustbox}
\end{center}
\end{table}

\vspace{-1.3cm}

\section{Conclusion} \label{conclusion}
In this paper, we provide a method to adapt classifiers from a source domain with a large category set to a target domain with few number of classes. The main contribution is a method to obtain an ordered set of  category specific source subspaces that match a particular target domain. The stopping criterion makes such a ordering practically useful (it could be a preprocessing step to enable a number of other domain adaptation techniques). Its use in the multi-source domain adaptation setting can serve widening the scope of domain adaptation to real world (`in the wild') settings where any combination of source domains can be made available.

In future, we would be interested in exploring domain adaptation in the wild setting for more challenging scenarios such as weakly supervised segmentation and in per sample domain adaptation settings.
\bibliographystyle{splncs}
\bibliography{egbib}
\end{document}